\documentclass[lettersize,journal]{IEEEtran}
\usepackage{amsmath,amsfonts}
\usepackage{algorithmic}
\usepackage{algorithm}
\usepackage{array}
\usepackage[caption=false,font=normalsize,labelfont=sf,textfont=sf]{subfig}
\usepackage{textcomp}
\usepackage{stfloats}
\usepackage{url}
\usepackage{verbatim}
\usepackage{graphicx}
\usepackage{cite}

\usepackage{multirow}
\usepackage{graphicx}
\usepackage{amsmath}
\usepackage{amssymb}
\usepackage{booktabs}
\usepackage{bbm}
\usepackage[table,xcdraw]{xcolor}

\def\eg{\emph{e.g.}} 

\def\ie{\emph{i.e.}}

\def\etal{\emph{et al.}}

\usepackage[utf8]{inputenc}

\begin{document}
\title{Suppressing Gradient Conflict for Generalizable Deepfake Detection}
\author{Ming-Hui Liu,~Harry Cheng,~Xin Luo,~Xin-Shun Xu*~\IEEEmembership{Senior Member, IEEE}
\thanks{*Corresponding author.

Ming-Hui Liu and Xin Luo are with the School of Software, Shandong University, Jinan 250101, China (e-mail: liuminghui@mail.sdu.edu.cn; luoxin.lxin@gmail.com).
Xin-Shun Xu is with the School of Software, Shandong University, Jinan 250100, China, and also with the Quan Cheng Laboratory, Jinan 28666, China (e-mail: xuxinshun@sdu.edu.cn).
Harry cheng is with the the School of Computer Science, National
University of Singapore, Singapore (e-mail: xaCheng1996@gmail.com).
}
}

\markboth{Journal of \LaTeX\ Class Files,~Vol.~14, No.~8, August~2021}%
{Shell \MakeLowercase{\textit{et al.}}: A Sample Article Using IEEEtran.cls for IEEE Journals}


\maketitle

\begin{abstract}
Robust deepfake detection models must be capable of generalizing to ever-evolving manipulation techniques beyond training data. A promising strategy is to augment the training data with online synthesized fake images containing broadly generalizable artifacts. 
However, in the context of deepfake detection, it is surprising that jointly training on both original and online synthesized forgeries may result in degraded performance. This contradicts the common belief that incorporating more source-domain data should enhance detection accuracy.
Through empirical analysis, we trace this degradation to gradient conflicts during backpropagation which force a trade-off between source domain accuracy and target domain generalization. To overcome this issue, we propose a Conflict-Suppressed Deepfake Detection (CS-DFD) framework that explicitly mitigates the gradient conflict via two synergistic modules. 
First, an Update Vector Search (UVS) module searches for an alternative update vector near the initial gradient vector to reconcile the disparities of the original and online synthesized forgeries. 
By further transforming the search process into an extremum optimization problem, UVS yields the uniquely update vector, which maximizes the simultaneous loss reductions for each data type. 
Second, a Conflict Gradient Reduction (CGR) module enforces a low-conflict feature embedding space through a novel Conflict Descent Loss. This loss penalizes misaligned gradient directions and guides the learning of representations with aligned, non-conflicting gradients. The synergy of UVS and CGR alleviates gradient interference in both parameter optimization and representation learning. Experiments on multiple deepfake benchmarks demonstrate that CS-DFD achieves state-of-the-art performance in both in-domain detection accuracy and cross-domain generalization.
\end{abstract}


\section{INTRODUCTION}
\begin{figure}
    \centering
    \includegraphics[width=0.48\textwidth]{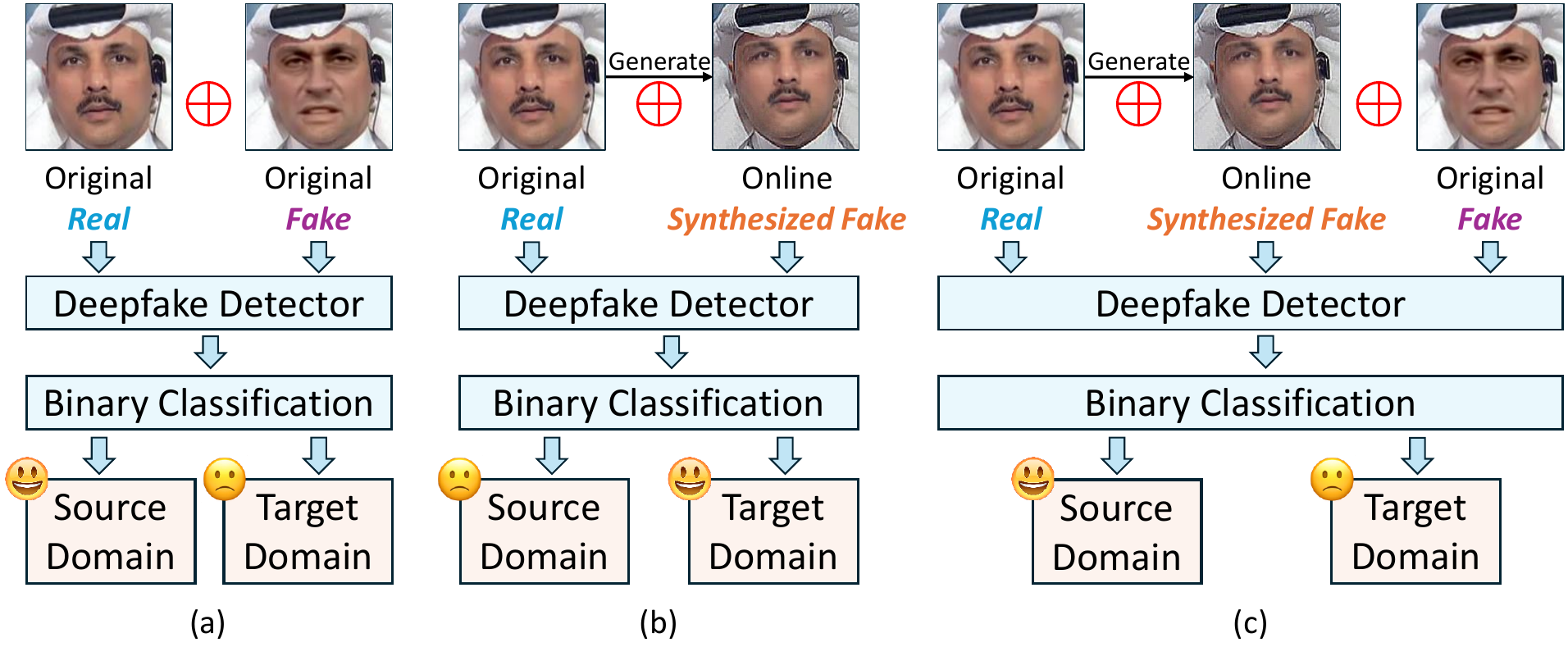}
    \caption{Different training strategies. (a) Traditional training strategy, which suffers from poor generalization. (b) Replacing original fake images with online synthesized ones, which lacks the ability to detect manipulations in the source domain. (c) Training with both online synthesized and original fake images, which struggles to generalize effectively to the target domain.}
    \label{fig:training_overview}
\end{figure}

\IEEEPARstart{W}{ith} the rapid development of multimodal large-scale models~\cite{SD-Latent-Diffusion, DDIM, DDPM, DCFace, DiFF}, the technical barriers to tampering or synthesizing facial data have been significantly lowered~\cite{Region_aware_swapping, DiffSwap, sun2025towards, DiffusionFace}. The proliferation of such deepfake content continues to disrupt the order of internet finance, social discourse, and ethical norms. 
Therefore, deepfake detection, which is dedicated to distinguishing authentic data from fake content, has gained attention from the research community~\cite{Prodet, Hong_Deepfake_CVPR_2024, xia2024mmnet, cvpr_2025_yan, Tan_CVPR24, sun2023contrastive, DF_PAMI2, Ensemble_1}. 
As illustrated in Fig.~\ref{fig:training_overview}(a), traditional deepfake detectors are trained with a fixed set of real and fake data to perform a binary classification~\cite{Xception, F3Net, MAT}. 
While this paradigm delivers strong performance under in-domain evaluation (where training and testing set share the same distribution), it often falters in cross-dataset settings, exhibiting markedly poorer generalization to unseen target domains~\cite{x-ray, SRM, RFM, CADDM}.

To enhance poor cross-domain generalization performance, a prominent approach is to replace original fake samples with online synthesized fake data generated from real images~\cite{SBI_ShioharaY22}.
These generated samples explicitly incorporate broadly generalizable artifacts rather than dataset‐specific cues~\cite{Freqblender}.
As illustrated in Fig.~\ref{fig:training_overview}(b), 
these newly synthesized fake samples are generated online~\cite{FCL+I2G}, and replace the original fake data in the training process. By disentangling the model from the artifacts of a fixed fake set, this strategy effectively improves the generalization ability of models in unseen domains.

Despite its benefits for cross-domain generalization, this training paradigm, \ie, replacing original fake examples with online synthesized fakes alongside real images, induces an disconcerting side effect: a significant degradation in source domain detection performance. 
Specifically, detectors trained in this way often fail to identify the forgeries contained in the original dataset. 
An intuitive remedy is to reintroduce the displaced original fake data. Nevertheless, as illustrated in Fig.~\ref{fig:training_overview}(c), doing so undermines the generalization ability in target domains.
This issue, also known as the `$1+1<2$' problem, reflects a counterintuitive phenomenon: More and richer source domain data do not necessarily lead to better detection performance, even though the online synthesized samples are generally considered a beneficial augmentation.
Several recent studies try to mitigate this challenge via superficial progressive training schemes~\cite{Prodet}. By contrast, our method takes a fundamentally different tack: We first investigate why merging these data sources can paradoxically impair performance, \ie, what underlying factors drive the `$1+1<2$' phenomenon?

\begin{figure}[t]
\centering
\includegraphics[width=0.48\textwidth]{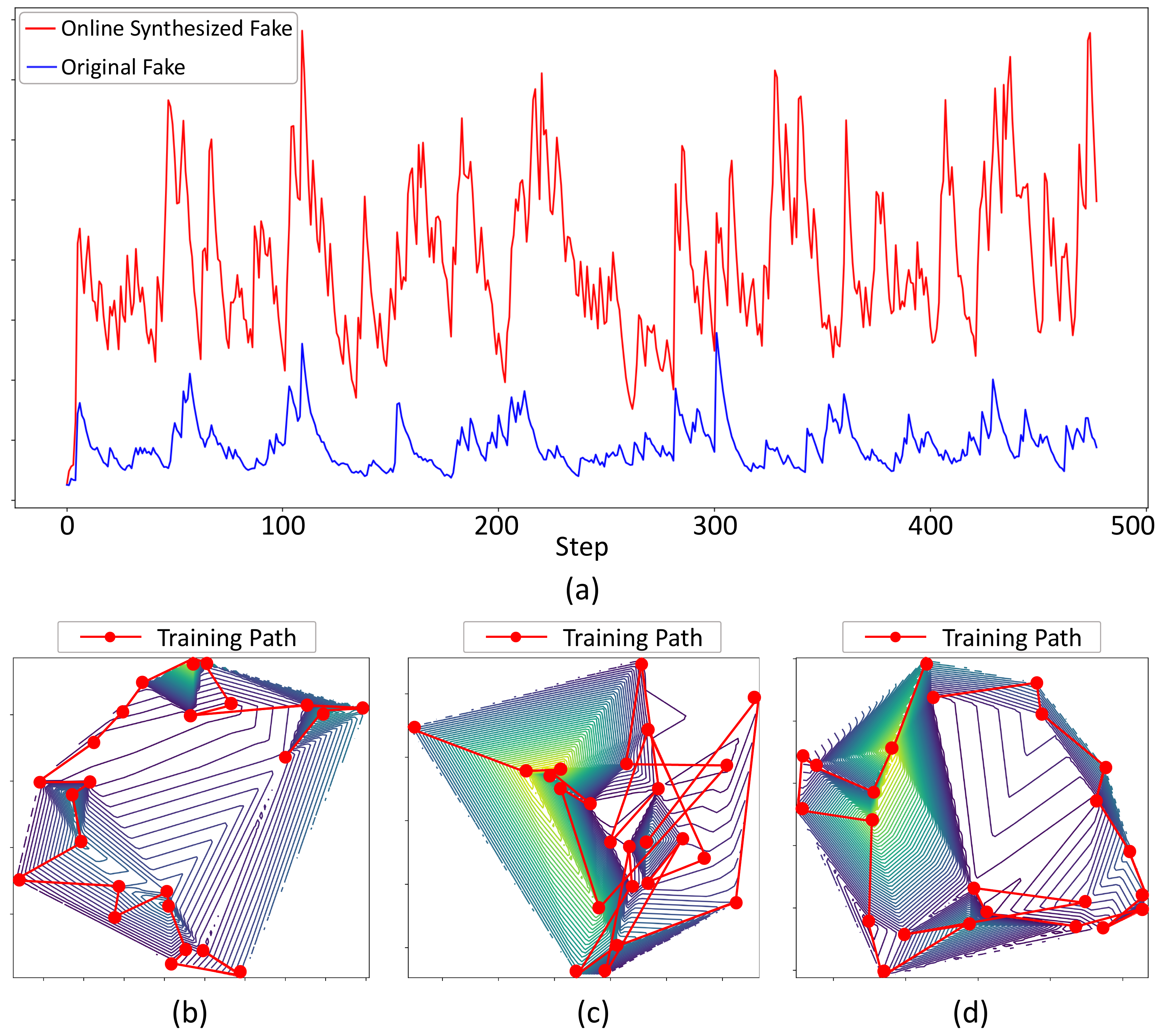}
\caption{Gradient conflicts when trained on different types of data. (a) When trained with two types of data, the model’s loss curve shows a fluctuating pattern with opposing peaks and valleys, indicating adversarial trends in optimization. (b) When trained on a single subset, the model converges smoothly. (c) However, when trained on both subsets simultaneously, the training becomes highly unstable, with significant oscillations. (d) Our CS-DFD method resolves the gradient conflict, presenting a smooth trend.}
\label{fig_intro}
\end{figure}

Our approach is motivated by intuitive comparisons of detector behavior on online synthesized versus original forgery data. To this end, we split the training set into two complementary subsets: \textbf{[real data, original fake data]} and \textbf{[real data, online synthesized fake data]}. 
During training, the detector concurrently processes both data subsets and computes the loss for each subset independently.
We then track the optimization trajectories and Fig.~\ref{fig_intro}(a) plots the classification losses for both training subsets over identical training iterations. From this figure, we identify a pronounced mutual suppression phenomenon: When one subset attains a lower loss value (located in the trough of the loss curve), the opposing subset exhibits a significantly higher loss (located in the peak of the loss curve). This observation implies that the model adopts contradictory learning logics to detect original forgery data and online synthesized forgery data, thereby deriving \textbf{\emph{conflicting gradient directions}}. 
To further validate this hypothesis, we visualize the loss landscape trajectories. As illustrated in Fig.~\ref{fig_intro}(b), when trained on a single subset,  the optimizer follows a smooth descent into a well‐defined local minimum.
However, as shown in Fig.~\ref{fig_intro}(c), when joint training on both subsets, the model becomes trapped in a saddle region near a high curvature gradient basin, exhibiting oscillatory behavior and failing to converge to the global optimum. The above divergence in convergence behavior underscores how the gradient conflicts of different subsets impede effective joint training.

Building on our empirical observations,  we pinpoint \emph{\textbf{gradient conflicts between heterogeneous forgery data}} as a fundamental barrier for addressing the `$1+1<2$' issue.
To achieve both high detection accuracy on the source domain and strong generalization to target domains, these conflicts must be explicitly resolved.
We thus propose Conflict-Suppressed Deepfake Detection (CS-DFD), a unified framework that harnesses discriminative cues from both original and online synthesized forgeries while ensuring stable convergence toward the global optimum. CS-DFD comprises two complementary modules: 
i) \emph{Update Vector Search (UVS).} 
Instead of applying the total gradient which oscillates between competing objectives, UVS computes the individual gradients and loss descent rates for both original subset and online synthesized subset. Then, it searches within the neighborhood of the total gradient to find a conflict-free update vector that maximizes the descent on both losses.
ii) \emph{Conflict Gradient Reduction (CGR).}
Recognizing that adjusting update vectors alone cannot fully eliminate gradient conflict, CGR introduces a conflict descent loss that quantifies and penalizes conflicting gradients in the embedding space. Thus, it i) encourages the network to learn a diverse set of discriminative features in shallow layers and ii) gradually maps these features into a low-conflict embedding space at deeper layers.
UVS and CGR allow CS-DFD to aggregate complementary features from all available forgery data without suffering the `$1+1<2$' degradation, thereby delivering both robust in-domain accuracy and superior cross-domain generalization. As shown in Fig.~\ref{fig_intro}(d), the optimization trajectory becomes noticeably smoother after applying our method. Our contributions are summarized as three-fold:

\begin{itemize}
\item{To the best of our knowledge, we are the first to identify the gradient conflicts inherent in heterogeneous forgery data and demonstrate that this issue induces a performance trade-off between the source and target domains.}
\item{To address the gradient conflicts, we propose a novel Conflict-Suppressed Deepfake Detection~(CS-DFD) framework which integrates two cooperative modules: The UVS module replaces the original gradient vectors using a conflict-free update vector, and the CGR module quantifies gradient conflict intensity and leverages this metric to learn a low-conflict embedding space.}
\item{We conduct comprehensive experiments on multiple deepfake datasets. And the results demonstrate that our method significantly outperforms several state-of-the-art baselines in both source and target domains. Moreover, our approach can seamlessly be integrated into different backbone architectures, exhibiting excellent versatility.}
\end{itemize}

\section{RELATED WORK}
\subsection{Deepfake Generation and Detection}
\noindent\textbf{Generation.} Benefiting from the continuous development of human portrait synthesis technologies, deepfake~\cite{DF_PAMI4} has increasingly become a prominent concern for communities~\cite{DF_PAMI1}. Early approaches in this field employ autoencoders~\cite{autoencoder} or generative adversarial networks (GANs)~\cite{GAN} to synthesize human likenesses~\cite{Face2Face}.
For instance, StyleGAN~\cite{StyleGAN} modifies high-level facial attributes with a progressive growing approach. IPGAN~\cite{IPGAN} disentangles the identity and attributes of the source and target faces, and then combines them to synthesize fake content. 
Moreover, identity-relevant features~\cite{3DMM2} have been introduced into deepfake generation~\cite{Region_aware_swapping}. 

In recent years, diffusion models~\cite{zhang2022gddim, Nichol_21_ICML, Lam_22_ICLR} have garnered attention for their ability to sample high-quality images by first adding noise to the original image and then progressively denoising it~\cite{Song_21_ICLR, Ho_et_al_DF_NIPS}. This adaptation of diffusion models into a foundation for deepfake synthesis has attracted significant interest. For instance, DiffSwap~\cite{DiffSwap} frames face swapping as a conditional inpainting problem by masking the target face and guiding the diffusion process with identity and landmark conditions. DiffFace~\cite{DiffFace} introduces an ID-conditional diffusion model and uses off-the-shelf facial expert models and a target-preserving blending strategy to transfer identities. DCFace~\cite{DCFace} presents a dual condition generator conditioned on both subject identity and external style factors to synthesize faces.

\noindent\textbf{Detection.} 
Deepfake detection~\cite{Hong_Deepfake_CVPR_2024,CVPR24_Yan_Aug,xia2024mmnet,guan2024improving, liu2025data} is generally cast as a binary classification task. Preliminary efforts often attempt to detect specific manipulation traces~\cite{Exploring_Frequency_Adversarial, SSTNET, pmlr-v235-zhang24aj}, which have shown certain improvements in the intra-data set setting. However, these methods often encounter inferior performance when applied to data with different distributions or manipulation methods. To address this generalization issue~\cite{Tan_CVPR24}, researchers have conducted investigations from multiple perspectives, such as more effective model architectures~\cite{Efficient, F3Net, ViT}, richer modal information~\cite{VFD, Leveraging_Real_Talking, emotion}, and more efficient data augmentation strategies~\cite{RFM, SRM, CADDM, DCT}. In recent years, online data synthesis methods have garnered significant attention~\cite{SBI_ShioharaY22, Self_ADV}. These methods highlight that fake samples in existing datasets often contain method-specific artifacts, which can lead to severe overfitting issues~\cite{cvpr_2025_yan, UCF_0002ZFW23, CVPR2025_1}. 
Therefore, these samples need to be replaced with more generalized fake images. For instance, Chen \etal~\cite{Self_ADV} propose a facial region blending strategy that generates fake samples with specified forged regions, blending types, and blending ratios. SBI~\cite{SBI_ShioharaY22} introduces a self-blending strategy, which mixes different real facial images during training to capture boundary-fusion features. These methods typically replace the original source-domain fake data with the online-synthesized fake images. However, they also introduce the `$1+1<2$' issue, where the detector is forced to trade off between detection performance in the source domain and generalization performance in the target domain.

\subsection{Gradient Conflict}
Historically, gradient conflict~\cite{liu2021conflict} issues have primarily been observed in multi-task learning and multi-objective optimization scenarios. It arises because models must simultaneously learn multiple unrelated tasks or optimize conflicting objectives. When different tasks or objectives require divergent—or even opposing—parameter update directions (\ie, the gradient vectors), significant gradient conflicts emerge. During training, conflicting gradients destabilize parameter updates, causing oscillatory behavior that impedes convergence or traps models in suboptimal solutions. Consequently, multi-task learning often results in performance degradation compared to individually training each task. To address this, PCGrad~\cite{yu2020gradient} indicates gradient conflict as the primary obstacle in multi-task learning and resolves it by projecting conflicting gradients onto their normal planes, ensuring stable convergence. GAC~\cite{le2024gradient} identifies ascending directions for each gradient to prevent any single gradient from dominating optimization or conflicting with others. SAGM~\cite{wang2023sharpness} minimizes empirical risk, perturbation loss, and their divergence simultaneously by maximizing gradient inner products. NASViT~\cite{gong2022nasvit} observes that gradients from different sub-networks conflict with those of the supernet and combine a gradient projection algorithm, a switchable layer scaling design, and a regularization training recipe to alleviate the conflict issue. However, these studies are confined to addressing gradient conflicts in traditional multi-task learning scenarios, failing to consider that such conflicts may also arise between different data within the same task. Concurrently, we are the first to recognize that gradient conflicts between different types of fake data in deepfake detection tasks severely impede the optimization process, thereby preventing existing methods from improving generalization by enhancing the diversity of fake data.
\section{PROBLEM FORMULATION}
The deepfake detection task can be formulated as a binary classification problem, optimized with the cross-entropy loss:
\begin{equation}
{\mathcal{L}_{\rm{bce}}} =  - \frac{1}{N}\sum\limits_{i = 1}^N {{y_i}\log {p_i}}  + \left( {1 - {y_i}} \right)\log \left( {1 - {p_i}} \right),
\label{eqn:bce}
\end{equation}
where $x_i$ denote the $i$-th input sample in the training set $X=[x_i]^N_{i=1}$, $N$ is the total number of samples. $y_i$ and $p_i$ are the ground-truth label and the predicted output for ${x}_i$, respectively. 
${X}$ serves as the source domain dataset, which consists of two subsets: real data ${x}_\text{r}$ and fake data ${x}_{\text{f}}$.
This training process achieves satisfactory detection performance on the source domain, while it has been widely recognized to induce overfitting, thereby posing generalization challenges on the target domain~\cite{x-ray}, \eg, other deepfake datasets.
To address this issue, a common approach is to replace original fake images ${x}_\text{f}$ with online synthesized ones ${x}_\text{s}$~\cite{FCL+I2G}. Specifically, the training input ${X} = [{x}_\text{r}, {x}_\text{f}]$ is substituted with ${X}' = [{x}_\text{r}, {x}_\text{s}]$. 
The motivation behind this replacement is to exploit prior knowledge to generate challenging samples that embody more generalizable artifacts, such as blending boundaries, thereby improving the detection ability across diverse domains. And this operation can be directly integrated into the training process described in Equation~(\ref{eqn:bce}).

However, due to the incompleteness of the training data~(absence of original forgery data ${x}_\text{f}$), models trained with ${X}'$ inevitably exhibit suboptimal performance on the source domain. To mitigate this limitation, a straightforward approach is to reintroduce ${x}_\text{f}$ as part of the training data to construct a new source domain training set,
\begin{equation}
    {X}'' = [{x}_\text{r}, {x}_\text{s}, {x}_\text{f}],
\end{equation}
where both ${x}_\text{f}$ and ${x}_\text{s}$ are the forgery data. While this strategy can partially restore performance on the source domain, it tends to compromise the generalization ability~\cite{Prodet}. Therefore, many studies implicitly assume an insurmountable inherent trade-off between source domain and target domain accuracy.

In this study, we aim to address the long-standing `$1+1<2$' problem from a novel perspective. Specifically, we reformulate the loss function in Equation~(\ref{eqn:bce}) into a dual-stream formulation, where the model simultaneously receives two types of input: ${X} = [{x}_\text{r}, {x}_\text{f}]$, \ie, real data and original forgeries, and ${X}' = [{x}_\text{r}, {x}_\text{s}]$, \ie, real data and online synthesized forgeries. For each input stream, we compute its loss function separately, and the total classification loss can be redefined as:
\begin{equation}
\begin{split}
\mathcal{L}_{\rm{bce}} &= {{\mathcal{L}_1} + {\mathcal{L}_2}}\\
&= { -\!\sum\limits_{j = 1}^2 {({\frac{1}{{{N_j}}}\!\sum\limits_{i = 1}^{{N_j}} {{y_i}\log {p_i}}\!+\!\left( {1 - {y_i}} \right)\log \left( {1 - {p_i}} \right)} )} },
\label{eqn:new_BCE}
\end{split}
\end{equation}
where $\mathcal{L}_1$ and $\mathcal{L}_2$ correspond to losses derived from different subsets. $j$ is the set index. Specifically, $\mathcal{L}_1$ is computed based on set ${X}$ and is considered effective for achieving strong performance on the source domain, while $\mathcal{L}_2$ is computed based on set ${X}'$ and can significantly enhance the generalization ability on target domains.
Equation~(\ref{eqn:new_BCE}) implies that, to achieve strong performance on both the source and target domains, $\mathcal{L}_1$ and $\mathcal{L}_2$ must be optimized cooperatively. In other words, at each training step $t$, the gradient vector $g_0$ is expected to drive the simultaneous minimization of both loss functions:
\begin{equation}
\begin{cases}
{{\mathcal{L}_{1,{\theta ^{t + 1}}}} < {\mathcal{L}_{1,{\theta ^t}}}}, \\ 
{{\mathcal{L}_{2,{\theta ^{t + 1}}}} < {\mathcal{L}_{2,{\theta ^t}}}}, 
\end{cases}
{\rm{s.t.}}~{\theta ^{t + 1}} = {\theta ^t} - \alpha {g_0},
\end{equation}
where $\alpha$ is a sufficiently small learning rate. $\theta^t$ denotes model parameters at time step $t$. 
However, as illustrated in Fig.~\ref{fig_intro}, gradient conflicts arise between the original forgery data and the online synthesized forgery data. These conflicts hinder effective cooperative optimization of the model, thereby giving rise to the aforementioned trade-off problem. That is, minimizing one loss will inevitably amplify the other:
\begin{equation}
{\mathcal{L}_{1,{\theta ^{t + 1}}}} > {\mathcal{L}_{1,{\theta ^t}}},~{\rm{s.t.}}~
\begin{cases}
{{\theta ^{t + 1}} = {\theta ^t} - \alpha {g_0}}, \\ 
{\mathcal{L}_{2,{\theta ^{t + 1}}}} > {\mathcal{L}_{2,{\theta ^t}}}.
\end{cases}
\label{eqn:gradient_conflict}
\end{equation}

To alleviate this challenge, we propose Conflict-Suppressed Deepfake Detection (CS-DFD) framework, which resolves such gradient conflicts to simultaneously enhance performance on both the source and target domains.
\section{METHODOLOGIES}
\begin{figure*}[t]

{\includegraphics[width=1\textwidth]{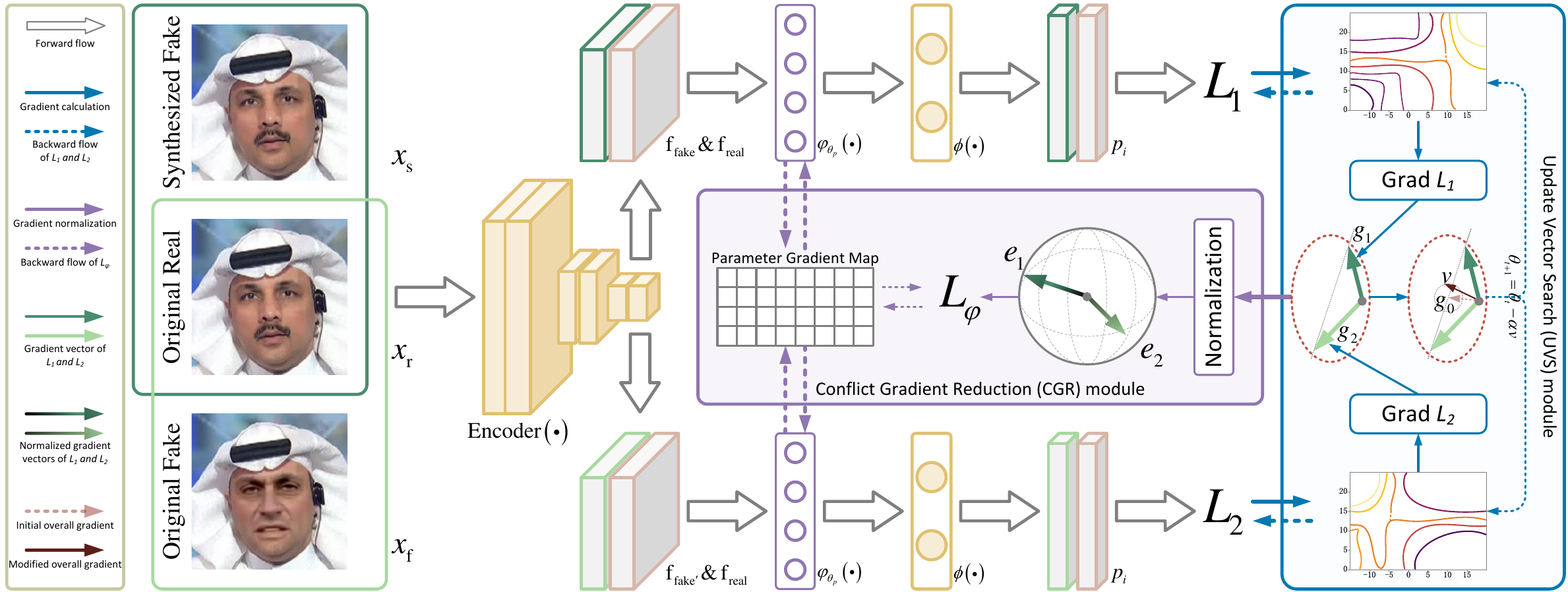}}

\caption{Overall architecture of the CS-DFD framework, which consists of two modules: Update Vector Search~(UVS) and  Conflict Gradient Reduction (CGR). After computing the gradient vectors $g_1$ and $g_2$ corresponding to two sets of data, UVS adjusts the overall gradient vector $g_0$ to obtain an approximate gradient vector $v$ that satisfies the descent requirements of loss functions $\mathcal{L}_1$ and $\mathcal{L}_2$; CGR measures the current level of conflict by $g_1$ and $g_2$, and leverages conflict descent loss $\mathcal{L}_{\varphi}(\cdot)$ to project the features into a low-conflict embedding space.}
\label{fig:framework}
\end{figure*}

The primary objective of our CS-DFD approach is to alleviate gradient conflicts between heterogeneous forgery data, thereby synchronously enhancing the detection accuracy on both source and target domains. As illustrated in Fig.~\ref{fig:framework}, our framework comprises two core modules: 1) the Update Vector Search~(UVS) module, which computes a conflict-free update vector to replace the original gradient vector; 2) the Conflict Gradient Reduction~(CGR) module, which quantifies conflict intensity using gradient vectors and leverages this measure to guide the model in learning a low-conflict embedding space. Both modules are designed to ensure the cooperative optimization of loss functions $\mathcal{L}_1$ and $\mathcal{L}_2$, thereby effectively reducing gradient conflict while ensuring that the capability of learning diverse knowledge from heterogeneous forgery data.

\subsection{Update Vector Search Module}
To simultaneously utilize both original forgeries and online synthesized forgeries while minimizing their corresponding losses $\mathcal{L}_1$ and $\mathcal{L}_2$, the UVS module searches for a conflict-free parameter update vector during the backpropagation stage. 

Usually, during the optimization process of a vanilla deepfake detector, its parameters are updated based on the gradient vector $g_0$ of the loss function $\mathcal{L}_{\rm{bce}}$. Thus, in each optimization iteration, the descent rate $\Delta_{\rm{bce}}$ of $\mathcal{L}_{\rm{bce}}$ is given by:
\begin{equation}
\begin{split}
\Delta_{\rm{bce}}  &= \frac{1}{\alpha }\left( {\mathcal{L}_{\rm{bce},\theta ^t} - \mathcal{L}_{\rm{bce},\theta ^{t+1}}}\right) \\
& = \frac{1}{\alpha }\left( {\mathcal{L}_{\rm{bce},\theta ^t} - \mathcal{L}_{{\rm{bce},\theta ^t} - \alpha {g_0}}} \right),
\end{split}
\end{equation}
where $\alpha$ is the learning rate; $\theta$ denotes the parameters.

However, as analyzed in previous Equation~(\ref{eqn:gradient_conflict}), loss components~($\mathcal{L}_1$ and $\mathcal{L}_2$) included in the total loss~($\mathcal{L}_{\rm{bce}}$) tend to counteract each other due to gradient conflicts. As a result, the descent of the total loss cannot guarantee the reduction in each individual loss component. When the overall gradient ${g}_0$ satisfies the descent requirement of one component, it may lead to an increase in the other.

To address this optimization dilemma, we try to search for a low-conflict update vector $v$ to replace the original gradient vector $g_0$ during the backpropagation process, thereby concurrently maximizing the reduction of both $\mathcal{L}_1$ and $\mathcal{L}_2$:
\begin{equation}
\begin{split}
\mathop {\max}\limits_{j \in[1,2]} {\Delta _j}  = 
\mathop {\max}\limits_{v \in \mathbb R} 
{\frac{1}{\alpha }( {{\mathcal{L}_{j,\theta}} - {\mathcal{L}_{j,{\theta - \alpha v}}}} )}.
\end{split}
\label{eqn:updated_vector}
\end{equation}
Considering that directly solving for a substitutional update vector $v$ is non-trivial, we apply a first-order Taylor expansion to Equation~(\ref{eqn:updated_vector}) and obtain the following simplified form:
\begin{equation}
\begin{split}
\mathop {\max}\limits_{j \in [1,2]} {\Delta _j} 
&\approx \mathop {\max}\limits_{v \in \mathbb R} 
{\frac{1}{\alpha }({\mathcal{L}_{j,\theta} - \mathcal{L}_{j,\theta} + \alpha {g_j}v})}  \\
&= {\max}~{g_j v},
\label{eq:max taylor}
\end{split}
\end{equation}
where $g_j$ denotes the gradient vector of the loss sub-term $\mathcal{L}_j$. 

In the context of resolving data conflicts, to ensure that both sub‐losses decrease in a coordinated manner, it is sufficient to maximizes the minimum descent rate of the two sub‐losses:
\begin{equation}
\begin{split}
\mathop {\max \min }\limits_{v \in \mathbb R,j \in [1,2]} {\Delta_j}
\!\approx\!\mathop {\max \min }\limits_{v \in \mathbb R,j \in [1,2]}{g_j v},
~{\rm{s.t.}}
\|{v - {g_0}}\|^2 \!\le\! c^2\| {g_0}\|^2,
\end{split}
\label{equ:neighbor}
\end{equation}
where the target update vector $v$ lies within the neighborhood $c\|{g_0}\|$ of the original gradient vector $g_0$ to ensure stability. In this way, the complex \textbf{set-based} problem has been transformed into a simpler \textbf{extremum-based} problem.

To further construct a solvable constrained linear optimization problem, we introduce an auxiliary variable $k$ to explicitly represent the lower bound of $\mathrm{min} {\Delta}_j$, and the original problem can be reformulated as solve for the maximum value of $k$:
\begin{equation}
\mathop{\max}\limits_{v \in \mathbb R, j \in [1,2]}~k, ~{\rm{s.t.}}~
\begin{cases}
{ {g_j v}  \ge k}, \\ 
{\|{v - {g_0}}\|^2 \le c^2\|{g_0}}\|^2.
\end{cases}
\label{eqn:linear_optimization_problem}
\end{equation}

To solve Equation~(\ref{eqn:linear_optimization_problem}), we construct its Lagrangian function to integrate Equation~(\ref{eqn:linear_optimization_problem}) with its constraint conditions according to the construction principles:
\begin{equation}
\begin{split}
\psi_{v,k,\lambda ,\mu }
\!=\! k\!-\!{\lambda}( {\|{v\!-\!{g_0}}\|^2 \!-\!c^2\|{g_0}\|^2}) 
\!-\! \sum\limits_{j = 1}^2 {{\mu _j}({k\!-\!{g_j v}})}.
\end{split}
\end{equation}

To ensure the existence of the maximum value of $k$, we impose $\sum\limits_{j = 1}^2 {{\mu _j} = 1}$. As a result, the influence of the variable $k$ can be eliminated, and the Lagrangian function simplifies to:
\begin{equation}
\begin{split}
\psi_{v,\lambda,\mu} = 
{\sum\limits_{j = 1}^2 {{\mu _j}{g_j v}} - {{\lambda}( {\|{v-{g_0}}\|^2 - c^2\|{g_0}\|^2})}}
\end{split}.
\end{equation}

At this point, primal problem is transformed into an optimization problem of the Lagrangian function by maximizing the variable $v$ and minimizing the variables $\lambda$ and $\mu$ (\ie, $\mathop {\max}\limits_{v}\mathop {\min}\limits_{\lambda,\mu} {\psi}$). Meanwhile, according to the strong duality principle, the optimal solution can be further approximated by minimizing the maximum of the Lagrangian function, \ie:
\begin{equation}
\begin{split}
\mathop{\min}\limits_{\lambda,\mu}\mathop{\max}\limits_{v}{\psi} = 
\mathop{\min}\limits_{\lambda,\mu}\mathop{\max}\limits_{v}
{{{g_w v}\!-\!{\lambda}({\|{v\!-\!{g_0}}\|^2 \!-\! c^2\|{g_0}\|^2})}   }
\end{split},
\label{eqn:Lagrangian_function}
\end{equation}
where $g_w = \sum\limits_{j = 1}^2 {{\mu _j}{g_j}}$. Subsequently, we analytically derive and solve for the stationary points with respect to the variables $v$ and $\lambda$, yielding the relevant extremum values $v^*$ and $\lambda^*$:
\begin{equation}
\begin{cases}
{v^ * } &= {g_0} + {g_w}/2{\lambda}, \\
{\lambda ^ * } &= {\|{g_w}\|}/2c\|{g_0}\|.
\end{cases}
\end{equation}

By substituting the obtained $v^*$ and $\lambda^*$ back, we derive the final dual objective, which involves only the variable $\mu$:
\begin{equation}
\mathop{\min}\limits_{\mu}{\psi}=\mathop{\min}\limits_{\mu}{{g_w}^T}{g_0} + c\|{g_0}\|\|{g_w}\|.
\end{equation}

Finally, we use the projected gradient method to minimize the final dual objective $\psi_\mu$ and obtain the optimal $\mu^*$. The optimization process is as follows:
\begin{equation}
\mu^{t+1}\leftarrow \mu^{t}-\beta\cdot\nabla_{\mu^t}\psi.
\end{equation}

After the above derivations, we uniquely determine the desired update vector $v^*$ and maximize the descent rate $\Delta_j$ of the subdominant sub-term. By seamlessly replacing the original gradient vector $g_0$ with a low-conflict update vector $v^*$ during backpropagation, the model can effectively maintain its detection performance in both the source and target domains.

\subsection{Conflict Gradient Reduction Module}
While the low-conflict update vector $v^*$ can simultaneously satisfy the descent requirements of both $\mathcal{L}_1$ and $\mathcal{L}_2$, it may increase the risk of losing specific discriminative knowledge associated with heterogeneous source forgery types. 
To preserve both common and specific knowledge from original and online synthesized forgeries, we insert a learnable feature projection layer ${\varphi_{{\theta _p}}}$ (with its parameters denoted as $\theta_p$) before the classifier $\phi(\cdot)$. In this way, features can be mapped into a low-conflict embedding space without affecting the preceding network. The projection operation is defined as:
\begin{equation}
\begin{split}
\hat{\mathbf f}  = {\varphi_{{\theta_p}}}({\mathbf f}),
\end{split}
\end{equation}
where $\mathbf f$ is the initial feature vector. $\hat{\mathbf f}$ is the output embedding that resides in a low-conflict space. Projection layer ${\varphi_{{\theta _p}}}$ is composed of several linear layers.

To obtain the desired projection layer ${\varphi_{{\theta _p}}}$, we should measure the intensity of gradient conflicts. As for the gradient vectors $g_1$ and $g_2$, which are computed from original fakes and online synthesized fakes, respectively, the intensity of their conflict can be quantified by the dot product of their unit vectors ${\mathbf e}_1$ and ${\mathbf e}_2$. Inspired by this, we propose a Conflict Descent Loss~$\mathcal{L}_{\varphi}(\cdot)$ to guide the reduction of the gradient conflict via ${\mathbf e}_1$ and ${\mathbf e}_2$:
\begin{equation}
\begin{split}
{\mathcal{L}_{\varphi}} 
= - {\mathbf e}_{1}^T \cdot {{\mathbf e}_{2}}
= - \frac{g_{1}^T}{\|{g_{1}}\|} \cdot \frac{g_{2}}{\|{g_{2}}\|}.
\end{split}
\end{equation}

It is worth noting that $\mathcal{L}_{\varphi}$ only affects the parameters $\theta_p$ in the projection layer $\varphi_{\theta_p}$. The preceding layers will not be impeded by the projection and can preserve as much specific knowledge as possible from heterogeneous forgery data.

Jointly considering the two binary classification loss (\ie, $\mathcal{L}_1$ and $\mathcal{L}_2$) and the conflict descent loss $\mathcal{L}_{\varphi}$, we formulate the total gradients of the feature projection layer $\varphi_{\theta_p}$ as follows:
\begin{equation}
\begin{split}
\nabla \mathcal{L}_{\theta _p} 
&= \nabla \mathcal{L}_{1} + \nabla \mathcal{L}_{2} +\nabla \mathcal{L}_{\varphi}\\
&= {g_{1,\theta _p}} + {g_{2,\theta _p}} - 
\frac{{H_{1,\theta _p}} \cdot {g_{2,\theta _p}} +
{H_{2,\theta _p}} \cdot {g_{1,\theta _p}}}{{\|{g_{1,\theta _p}}\|}{\|{g_{2,\theta _p}}\|}},
\end{split}
\end{equation}
where $H$ is the Hessian matrix, representing the second-order partial derivative of the corresponding loss function $\mathcal{L}_{\varphi}(\cdot)$. Given the parameters $\theta_p=[\theta_1, \theta_2, \dots, \theta_n]$ of projection layer $\varphi_{\theta_p}$, $H$ can be mathematically defined as:
\begin{equation}
H = \nabla _{\theta_p} ^2\mathcal{L}_{\varphi} = \left[ {\begin{array}{*{20}{c}}
{\frac{{{\partial ^2}\mathcal{L}_{\varphi}}}{{\partial \theta _1^2}}}&{\frac{{{\partial ^2}\mathcal{L}_{\varphi}}}{{\partial {\theta _1}{\theta _2}}}}& \cdots &{\frac{{{\partial ^2}\mathcal{L}_{\varphi}}}{{\partial {\theta _1}{\theta _n}}}}\\
 \cdots & \cdots & \ddots & \cdots \\
{\frac{{{\partial ^2}\mathcal{L}_{\varphi}}}{{\partial {\theta _n}{\theta _1}}}}&{\frac{{{\partial ^2}\mathcal{L}_{\varphi}}}{{\partial {\theta _n}{\theta _2}}}}& \cdots &{\frac{{{\partial ^2}\mathcal{L}_{\varphi}}}{{\partial \theta _n^2}}}
\end{array}} \right].
\end{equation}

Since directly computing the full Hessian matrix $H$ is computationally infeasible, we resort to finding an approximation. Given the properties of the Fisher information matrix~(FIM), it is reasonable to assume that, under asymptotic conditions, the Hessian matrix $H$ can be well approximated by the FIM, which is typically estimated via the gradient outer product matrix $G$ during practical training process:
\begin{equation}
\begin{split}
H \approx {\rm {FIM}}=\mathbb{E}_{y|x}G.
\end{split}
\end{equation}

For a given gradient vector $g\!\in\!\mathbb R^{n}$, its corresponding gradient outer product matrix $G\in \mathbb R^{n \times n}$ is formulated as:
\begin{equation}
G = {g} \cdot {g}^T = \left[ {\begin{array}{*{20}{c}}
{{g}_1^2}&{{{g}_1}{{g}_2}}& \cdots &{{{g}_1}{{g}_n}}\\
 \cdots & \cdots & \ddots & \cdots \\
{{{g}_n}{{g}_1}}&{{{g}_n}{{g}_2}}& \cdots &{{g}_n^2}
\end{array}} \right].
\end{equation}

To further reduce computational overhead, we diagonalize the outer product matrix $G$, and finally approximate $H$ as:
\begin{equation}
\begin{split}
H \approx \tau \cdot{\rm{Diag}}( G ) = \tau  \cdot {g} \otimes {g},
\label{eq:g*g}
\end{split}
\end{equation}
where the hyper-parameter $\tau$ is used to control the variance. The operator $\otimes$ denotes element-wise (Hadamard) product. 

Ultimately, according to Equation~(\ref{eq:g*g}), the gradient of the feature projection layer $\varphi_{\theta_p}$ can be calculated as follows:
\begin{equation}
\begin{split}
{\nabla {\mathcal{L}_{\theta _p}}}
&= {g_{1,\theta}}+{g_{2,\theta}} \\
&- \gamma ({{g}_{1,\theta_p}}\!\otimes\!{{g}_{1,\theta_p}}\!\otimes\!{{g}_{2,\theta_p}} +{{g}_{2,\theta_p}}\!\otimes\!{{g}_{2,\theta_p}}\!\otimes\!{{g}_{1,\theta_p}}), \\
\end{split}
\end{equation}
where denote $\gamma = \frac{\tau}{\|g_{1,\theta_p}\|\|g_{2,\theta_p}\|}$. Guided by the conflict descent loss $\mathcal{L}_{\varphi}$, the feature projection layer $\varphi_{\theta_p}$ gradually maps features into a low-conflict embedding space, while preserving the learned specific patterns of heterogeneous forgery types.

Overall, our proposed CS-DFD primarily focuses on the computation and utilization of gradient vectors. Specifically, the UVS module computes a low-conflict approximate gradient, while the CGR module leverages the gradients to facilitate the learning of low-conflict representations. Through the collaborative effect of these two modules, our approach effectively eliminates gradient conflicts among heterogeneous forgery data, while preserving detection performance in the source domains and enhancing generalization in the target domains. The workflow of our method is outlined in Algorithm~\ref{alg:alg1}.
\begin{algorithm}[t]
\caption{An overview of our proposed CS-DFD.}\label{alg:alg1}
\begin{algorithmic}
\STATE 
\STATE {\textsc{\textbf{TRAIN}}}
\STATE $\textbf{For $i$ to MaxEpoch do}$
\STATE \hspace{0.5cm} $\textrm{\textbf{Input:} $N$ labeled samples $[x_i,y_i]^N_{i=1}$} $
\STATE \hspace{0.5cm} $\textrm{from two subsets ${X} = [{x}_\text{r}, {x}_\text{f}]$ and ${X}' = [{x}_\text{r}, {x}_\text{s}]$}$;
\STATE \hspace{0.5cm} $\textrm{\textbf{Step 1:} Feature extraction and projection;}$
\STATE \hspace{0.5cm} $\hat{\mathbf f}  = {\varphi_{{\theta_p}}}({\mathbf f})$;
\STATE \hspace{0.5cm} $\textrm{\textbf{Step 2:} Gradient calculation;}$
\STATE \hspace{0.5cm} $\nabla \mathcal{L}_{\theta _p}=\nabla \mathcal{L}_{1} + \nabla \mathcal{L}_{2} +\nabla {\mathcal L}_{\varphi}$;
\STATE \hspace{0.5cm} $\nabla \mathcal{L}_{\varphi} \approx - \gamma ({{g}_{1,\theta_p}}\otimes{{g}_{1,\theta_p}}\otimes{{g}_{2,\theta_p}} +{{g}_{2,\theta_p}}\otimes{{g}_{2,\theta_p}}\otimes{{g}_{1,\theta_p}})$;
\STATE \hspace{0.5cm} $\textrm{\textbf{Step 3:} Gradient modification;}$
\STATE \hspace{0.5cm} ${g_0} \gets {v^*} = {g_0} + (\sum\limits_{j=1}^2 {{\mu^*_j}{g_j}})/2{\lambda}$;
\STATE \hspace{0.5cm} ${\mu^*} \gets \mu^{t+1} = \mu^{t} - \beta\cdot\nabla_{\mu^t}\psi$;
\STATE \hspace{0.5cm} $\textrm{\textbf{Step 4:} Gradient backpropagation;}$
\STATE \hspace{0.5cm} $\theta^{t+1}=\theta^{t} - \alpha\cdot v^*$;
\STATE \hspace{0.5cm} $\theta_{p+1}^{t+1}=\theta_p^{t} - \alpha\cdot \nabla {\mathcal L}_{\theta _p}$;
\STATE $\textbf{End}$

\STATE {\textsc{\textbf{PREDICT}}}
\STATE \hspace{0.5cm} $\textrm{\textbf{Input:} $N'$ samples from difference target domains;} $
\STATE \hspace{0.5cm} $\textrm{\textbf{Step 1:} Feature extraction and projection;}$
\STATE \hspace{0.5cm} $\textrm{\textbf{Step 2:} Classification;}$
\end{algorithmic}
\label{alg1}
\end{algorithm}

\section{EXPERIMENTS}
\subsection{Implementation}
\noindent\textbf{Training datasets.} Following the common setting of deepfake detection studies~\cite{CVPR24_Yan_Aug,guan2024improving, liu2025data, Exploring_Frequency_Adversarial, SSTNET, pmlr-v235-zhang24aj}, we trained our model on the FF++ dataset~\cite{Xception}. It consists of 1,000 real videos and 5,000 manipulated videos generated using five different forgery methods, \ie, Deepfakes (DF), FaceSwap (FS), Face2Face (F2F), FaceShifter (Fsh), and NeuralTextures (NT), resulting in a total of 6,000 videos. 

\noindent\textbf{Testing datasets.} 
The official test split of the FF++ dataset is used to evaluate the detection performance on the source domain. Furthermore, four widely used deepfake datasets represent the target domain to evaluate the generalizability: \noindent\textbf{1) Celeb-DF}~\cite{Celeb-DF} is one of the most challenging datasets for deepfake detection, which contains 590 original YouTube videos and 5,639 corresponding high-quality deepfake videos. 

\noindent\textbf{2) DFDC}~\cite{dfdc} is one of the largest public deepfake datasets by far, with 23,654 real videos and 104,500 fake videos generated by eight facial manipulation methods. 
\noindent\textbf{3) DFDCp}~\cite{dfdc} offers a smaller subset of 1,131 real videos and 4,113 forged videos, intended for rapid prototyping and preliminary evaluation ahead of the full DFDC.
\noindent\textbf{4) UADFV}~\cite{li2018ictu} consists of 98 videos (49 real and 49 fake) created using FakeApp.

We utilize Dlib\footnote{http://dlib.net/.} to extract faces and resize them to $256 \times 256$ pixels for both the training and testing sets. Our experiments are conducted on a single RTX 3090 GPU with a batch size of 16, and the EfficientNet~\cite{Efficient} is employed as the backbone for CS-DFD. Also, to demonstrate the robustness of CS-DFD, we switch the backbone to other networks, \textit{e.g.}, ViT~\cite{ViT} in Section~\ref{sec:exp_backbone}.
The hyperparameters $c$ in the UVS module and $\tau$ in the CGR module are 0.5 and 0.01, respectively. A detailed analysis of these choices can be found in Section~\ref{sec:hyper}.

\subsection{Performance Comparison}

\begin{table*}[t]
\caption{Performance comparison (\%). 
The `Data Usage' column indicates the type of forged data utilized by the model. Specifically, `Original' denotes originally fake forgeries, whereas `Online' refers to online forgeries.
$^{\ddag}$: We re-implemented this detector. -: The authors did not report the results on this dataset in their original paper.}
\label{Tab:SOTA}
\centering
\begin{tabular}{l|r|cc|cccccc}
\toprule \midrule

\multicolumn{1}{l|}{\multirow{2}{*}{Method}} & \multicolumn{1}{c|}{\multirow{2}{*}{Venue}} & \multicolumn{2}{c|}{Data Usage}     & \multicolumn{6}{c}{Testing Dataset}                                                                                                                              \\ \cmidrule{3-10} 
\multicolumn{1}{l|}{}                        & \multicolumn{1}{c|}{}                       & Original             & \multicolumn{1}{c|}{Online} & \multicolumn{1}{c|}{FF++}                 & \multicolumn{1}{c}{Celeb-DF} & \multicolumn{1}{c}{DFDC} & \multicolumn{1}{c}{DFDCp} & \multicolumn{1}{c}{UAVDF} & \multicolumn{1}{c}{AVG} \\ \cmidrule(lr){1-2} \cmidrule(lr){3-3} \cmidrule(lr){4-4} \cmidrule(lr){5-5} \cmidrule(lr){6-6} \cmidrule(lr){7-7} \cmidrule(lr){8-8}  \cmidrule(lr){9-9}    \cmidrule(lr){10-10} 
 
\multicolumn{1}{l|}{$^{\ddag}$EfficientNet~\cite{Efficient}}    & ICML'19   &$\checkmark$  & $\times$
& 96.67 & \multicolumn{1}{|c}{64.59}    &  65.43   &  80.27 & 63.19  & \multicolumn{1}{c}{68.37}       \\
\multicolumn{1}{l|}{$^{\ddag}$Face X-ray~\cite{x-ray}}                     & CVPR'20     & $\checkmark$ & $\times$
& 95.72 & \multicolumn{1}{|c}{74.76}  &  61.57   & 71.15  & 64.34  & \multicolumn{1}{c}{67.95}       \\
\multicolumn{1}{l|}{$^{\ddag}$CORE~\cite{CORE_CVPRW_2022}}   &CVPRW'22    & $\checkmark$ & $\times$
& 96.61 & \multicolumn{1}{|c}{79.45}  &  62.60   & 75.74  &  65.41 & \multicolumn{1}{c}{70.80}       \\
\multicolumn{1}{l|}{$^{\ddag}$RECCE~\cite{Face_Reconstruction}}   &CVPR'22   & $\checkmark$ & $\times$
& 96.95 & \multicolumn{1}{|c}{69.71}   & 62.82  & 74.19  & 78.61  & \multicolumn{1}{c}{71.33}       \\
\multicolumn{1}{l|}{$^{\ddag}$UCF~\cite{UCF_0002ZFW23}}           & ICCV'23     & $\checkmark$ &  $\times$    
& 97.16 & \multicolumn{1}{|c}{81.90}   & 66.21  &  80.94 & 93.15  & \multicolumn{1}{c}{80.55}       \\
\multicolumn{1}{l|}{FoCus~\cite{tian2024learning}}      & TIFS'24          & $\checkmark$ & $\times$
& \textbf{99.15} & \multicolumn{1}{|c}{76.13}    &68.42&76.62& -  & \multicolumn{1}{c}{-}       \\
\multicolumn{1}{l|}{Qiao et al.~\cite{qiao2024fully}}   & TPAMI'24        & $\checkmark$ &   $\times$
& 99.00 & \multicolumn{1}{|c}{70.00} &-    &-    &78.00& \multicolumn{1}{c}{-}       \\
\multicolumn{1}{l|}{GRU~\cite{choi2024exploiting}}      & CVPR'24        & $\checkmark$ &   $\times$
& 98.40 & \multicolumn{1}{|c}{89.00}   &  -   & -  & -  & \multicolumn{1}{c}{-}       \\ 
\multicolumn{1}{l|}{$^{\ddag}$Effort~\cite{Effort}}      & ICML'25       & $\checkmark$ &   $\times$ 
& 98.85 & \multicolumn{1}{|c}{94.38}   &  74.49   & 84.22  & 95.07  & \multicolumn{1}{c}{87.04}       \\ 
\multicolumn{1}{l|}{$^{\ddag}$VLFFD~\cite{VLFFD}}      & CVPR'25          & $\checkmark$ & $\times$
& 98.64 & \multicolumn{1}{|c}{84.80}   &  71.80   & 84.74  & 93.71  & \multicolumn{1}{c}{83.76}       \\ 
\midrule
\multicolumn{1}{l|}{$^{\ddag}$I2G-PCL~\cite{FCL+I2G}}         & ICCV'21     & $\times$  & $\checkmark$       
& 92.21 & \multicolumn{1}{|c}{71.12} &65.55  & 73.58  & 94.08 & \multicolumn{1}{c}{76.08}       \\
\multicolumn{1}{l|}{$^{\ddag}$SBI~\cite{SBI_ShioharaY22}}         & CVPR'22     & $\times$  & $\checkmark$       
& 85.16 & \multicolumn{1}{|c}{93.18} &72.42  & 84.15  & 94.28 & \multicolumn{1}{c}{86.00}       \\
\multicolumn{1}{l|}{$^{\ddag}$FreqBlender~\cite{Freqblender}}             &   NeurIPS’24     & $\times$  & $\checkmark$
& 93.29 & \multicolumn{1}{|c}{92.65}   & 73.15  & 84.56  & 94.79  & \multicolumn{1}{c}{86.28}       \\
\multicolumn{1}{l|}{ProDet~\cite{Prodet}}             &   NeurIPS’24   & $\checkmark$ & $\checkmark$
& 95.91 & \multicolumn{1}{|c}{84.48}    &  72.40  & 81.16  & -  & \multicolumn{1}{c}{-}       \\
\midrule  
\rowcolor[HTML]{D9EEF2}\multicolumn{1}{l|}{CS-DFD} & \multicolumn{1}{c|}{-}      & $\checkmark$ & $\checkmark$
&{98.88} & \multicolumn{1}{|c}{\textbf{95.32}} &\textbf{74.97} &\textbf{85.86}  &\textbf{95.92}  & \multicolumn{1}{c}{\textbf{88.02}}       \\
\midrule 
\bottomrule
\end{tabular}
\end{table*}

\subsubsection{Comparison with SoTA detectors}
We present the comparison results between our proposed method CS-DFD and SoTA models in Table~\ref{Tab:SOTA}. 
All of them are trained on FF++ and evaluated on five testing datasets, covering both the source and target domains. We use the Area Under the Curve (AUC) metric to evaluate performance on each individual dataset and also report the average AUC across the four target-domain test sets.
From Table~\ref{Tab:SOTA}, we have the following observations: 

i) Training the model solely on original forgeries preserves high accuracy in the source domain, but fails to ensure generalization to the target domain. For example, although FoCus achieves the highest source-domain accuracy of 99.15\% due to overfitting behavior, its target-domain AUC drops to 76.13\% on Celeb-DF, highlighting poor generalization and limiting its effectiveness in practical applications.

ii) Training the model solely on online synthesized images yields the opposite results. For example, SBI achieves an average AUC of 86\% across four cross-domain datasets, substantially outperforming other baselines, but it suffers from a pronounced drop in performance on the source domain, achieving an AUC of 85\% on FF++. 

iii) Training on both types of forgeries, our method achieves superior performance on the FF++ dataset as well as across all cross-dataset evaluations. For instance, our method achieves an AUC of about 99\% on the FF++ dataset, surpassing all baselines that rely on online synthesized images, and performs on par with traditional detectors. Moreover, our approach substantially outperforms all baseline methods, achieving an average AUC of 88\% across the target-domain testing datasets.

The above results demonstrate that our CS-DFD overcomes the gradient conflicts between heterogeneous forgery data, thereby effectively leveraging more diverse training data to enhance model performance.

\begin{table}[t]
\centering
\caption{Performance of applying CS-DFD to different backbones. ViT-L and Vit-B represent different scales of the ViT.}
\scalebox{0.8}{
\begin{tabular}{l|cccccc}
\toprule \midrule
\multirow{1}{*}{Backbone}           
& \multicolumn{1}{|c}{FF++} & \multicolumn{1}{|c}{Celeb-DF} & \multicolumn{1}{c}{DFDC} & \multicolumn{1}{c}{DFDCp} & \multicolumn{1}{c}{UADFV} \\ \cmidrule(lr){1-1} \cmidrule(lr){2-2} \cmidrule(lr){3-3} \cmidrule(lr){4-4} \cmidrule(lr){5-5} \cmidrule(lr){6-6}
\multicolumn{1}{l|}{Xception (O)}               
&96.37 & \multicolumn{1}{|c}{56.75}   & 64.19 & 72.17 &62.05  \\ 
\multicolumn{1}{l|}{Xception (S)}               
&77.50 & \multicolumn{1}{|c}{90.11}   &70.16  &73.19  &94.12     \\
\multicolumn{1}{l|}{Xception (O+S)}               
&94.86 & \multicolumn{1}{|c}{87.77}   &68.96  &72.36  &92.87     \\
\multicolumn{1}{l|}{Xception (CS-DFD)}               
&\textbf{97.48} & \multicolumn{1}{|c}{\textbf{88.44}}   & \textbf{74.01} & \textbf{79.89} & \textbf{95.16}      \\ \midrule
\multicolumn{1}{l|}{ViT-L (O)}                         
&97.65 & \multicolumn{1}{|c}{77.27}  &64.54  & 78.05  & 76.13   \\ 
\multicolumn{1}{l|}{ViT-L (S)}                         
&70.32 & \multicolumn{1}{|c}{88.26}  &71.98  &84.60   &93.71    \\
\multicolumn{1}{l|}{ViT-L (O+S)}                         
&90.66 & \multicolumn{1}{|c}{85.15}  &70.02  &83.88   &91.34    \\
\multicolumn{1}{l|}{ViT-L (CS-DFD)}                         
&\textbf{97.89} & \multicolumn{1}{|c}{\textbf{90.26}}  &\textbf{73.83}  &\textbf{87.15}   &\textbf{94.92}    \\ \midrule
\multicolumn{1}{l|}{ViT-B (O)}                          
&98.38 & \multicolumn{1}{|c}{89.63}  &68.00  & 80.04  & 80.17    \\
\multicolumn{1}{l|}{ViT-B (S)}                          
&89.64 & \multicolumn{1}{|c}{93.54}  &73.63  & 84.87   &92.04     \\
\multicolumn{1}{l|}{ViT-B (O+S)}                          
&98.17& \multicolumn{1}{|c}{91.28}   &71.11  & 82.68   &90.21     \\
\multicolumn{1}{l|}{ViT-B (CS-DFD)}                          
&\textbf{98.44} & \multicolumn{1}{|c}{\textbf{93.66}}  &\textbf{76.92}  & \textbf{85.30}  &\textbf{94.85}   \\
\midrule 
\bottomrule
\end{tabular}
}
\label{Tab:backbone}
\end{table}

\subsubsection{Comparison with different backbones}
\label{sec:exp_backbone}
To validate the generalizability of our CS-DFD, we apply it to multiple backbone architectures, \ie, Xception, ViT-L, and ViT-B. For each backbone, we conduct training under three baseline strategies: i) original real + original forgeries (denoted as \textbf{(O)}), ii) original real + online synthesized images (denoted as \textbf{(S)}), and iii) original real + original forgeries + online synthesized images (denoted as \textbf{(O+S)}). We compare the performance of these baselines with their counterparts enhanced by our CS-DFD method and summarized the results in Table~\ref{Tab:backbone}.

Across all three backbone architectures, we consistently observe that incorporating online synthesized images during training significantly degrades detection performance in the source domain. For instance, ViT-L trained with synthesized forgeries (ViT-L (S)) achieves only 70\% AUC on FF++, a notable drop from the 97\% AUC when trained exclusively on original forged images (ViT-L (O)). Moreover, a counterintuitive `$1+1 < 2$' phenomenon arises: combining both original and online synthesized forgeries during training leads to performance degradation on both the source and target domains. For instance, Xception trained on the combined dataset (Xception (O+S)) achieves approximately 87\% AUC on Celeb-DF, which is significantly lower than the 90\% AUC achieved by Xception (S).
In contrast, our proposed CS-DFD method effectively mitigates these limitations. For example, when applied to ViT-B, CS-DFD enables the model to surpass all baseline strategies across all datasets. This improved generalization performance on both source and target domains stems from CS-DFD resolving gradient conflicts by balancing the optimization trajectory for the two conflicting data types.

\subsection{Ablation Studies}
\label{sec:exp_abl}
\begin{table}[t]
\centering
\caption{AUC (\%) comparison of different modules in CS-DFD.}
\scalebox{0.80}{
\begin{tabular}{ccc|ccccc}
\toprule
\midrule
\multirow{2}{*}{Backbone} & \multirow{2}{*}{UVS}   &\multirow{2}{*}{CGR}
&\multicolumn{4}{c}{Testing Set}                   \\ \cmidrule{4-8} 
& & & FF++ &Celeb-DF &DFDC &DFDCp &UADFV                                \\ \midrule
\checkmark  & &  
& 96.67 &{64.59} &{65.43} &{80.27} &{63.19} \\
\checkmark &\checkmark  & 
&{97.98}&{94.45}&{73.65}&{84.73}&{93.32} \\ 
\checkmark &  &\checkmark
& 98.00 &{94.92}&{74.64}&{84.44}&{94.09} \\ 
\checkmark &\checkmark &  \checkmark
&\textbf{98.88} &\textbf{95.32} &\textbf{74.97} &\textbf{85.86}  &\textbf{95.92} \\
 \midrule \bottomrule
\end{tabular}
}
\label{tab:ablation}
\end{table}
We report the ablation study results of our proposed method in Table~\ref{tab:ablation}. Specifically, we progressively integrate the two proposed modules, UVS and CGR, into the backbone model to examine their individual and combined contributions to performance. The results show that both modules independently contribute positively to model performance.
For instance, the UVS module improves the AUC on Celeb-DF by 30\%. Similarly, the CGR module yields a 9\% AUC improvement on the DFDC dataset. It is worth noting that throughout the ablation experiments, our method does not suffer performance degradation in the source domain, even using a combination of online synthesized and original forgery images. This indicates that both modules are effective in alleviating gradient conflict.
By combining UVS and CGR, the model achieves the best overall performance, confirming the complementary strengths of the two modules and their joint effectiveness in addressing training conflicts and improving cross-domain generalization.

\subsection{Visualization}

\begin{figure}[t]
\centering
\includegraphics[width=0.48\textwidth]{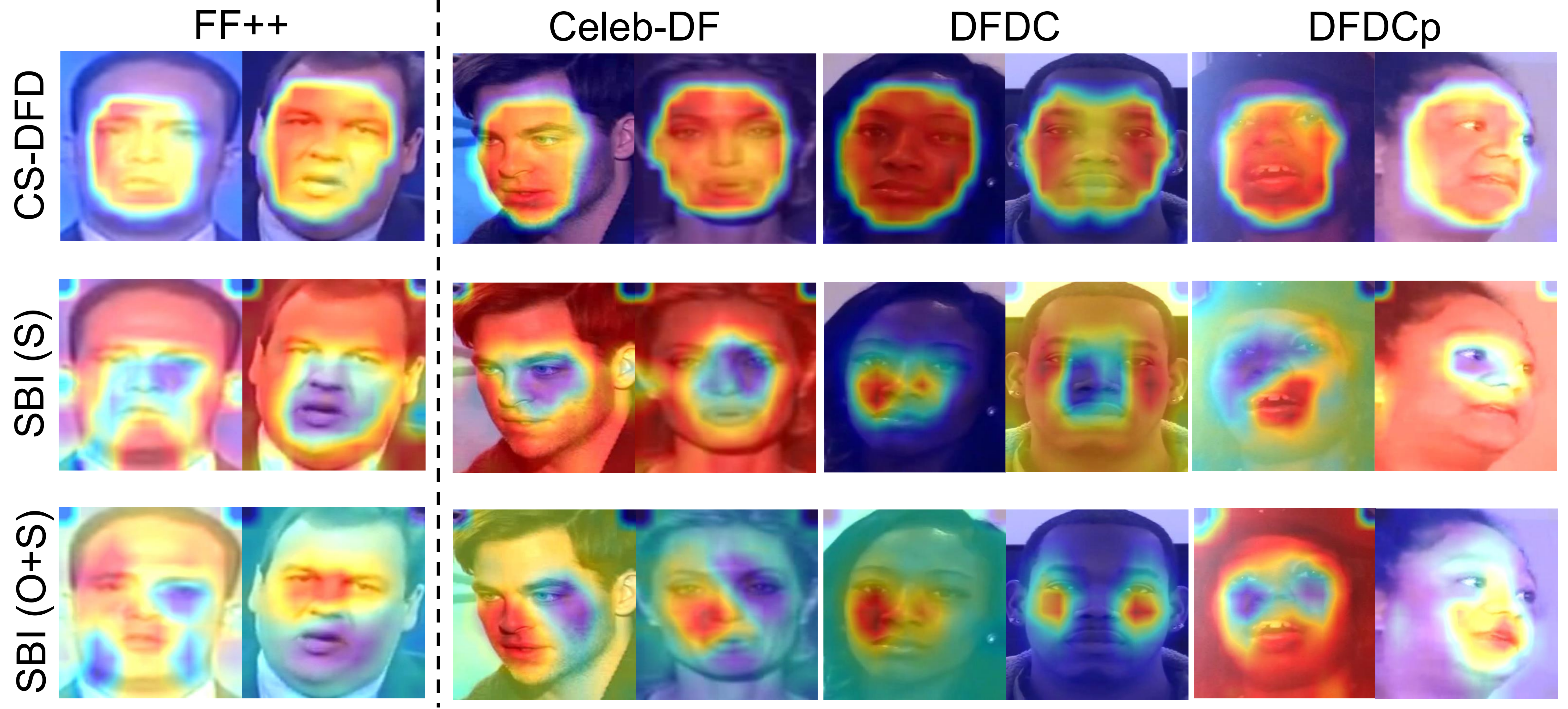}
\caption{The heatmaps of three different training strategies (\ie, CS-DFD, SBI (S) and SBI (O+S)) on source domain dataset (\ie, FF++) and target domain datasets (\ie, Celeb-DF, DFDC, and DFDCp). The intensity of red color corresponds to the level of attention.}
\label{fig_cam}
\end{figure}

\subsubsection{Heat Maps}
Fig.~\ref{fig_cam} presents the heatmaps generated by three different methods, \ie, CS-DFD, SBI (S), and SBI (O+S). Among them, SBI (S) is trained with \textbf{[real, online synthesized fake images]}, whereas SBI (O+S) and CS-DFD are trained with \textbf{[real, online synthesized fake images, original fake images]}. As observed in Fig.~\ref{fig_cam}, SBI (S) exhibits a consistent detection tendency across all target domain fake data, primarily focusing on the contours of facial blending. This consistent behavior contributes to improved generalization performance. However, SBI’s ability to capture more subtle forgery artifacts is limited. For instance, the heatmap in SBI (S) shows that the model tends to overlook the central facial region, thereby reducing its in-domain effectiveness. In contrast, SBI (O+S) exhibits confused detection patterns across the fake images. Specifically, its attention regions exhibit considerable uncertainty — for instance, on DFDCp, the model sometimes focuses heavily on the eyes, while at other times it attends to the lips. This confusion arises from gradient conflicts introduced by the inclusion of heterogeneous data during training, which ultimately disrupts the optimization direction and degrades performance. 
In contrast, our method, CS-DFD, clearly overcomes these issues. For instance, it consistently exhibits stable attention regions across different domains, with broader focus areas and richer forensic features. This enables the model to achieve consistently high accuracy across both in-domain and out-of-domain scenarios.

\begin{figure}[t]
\centering
\includegraphics[width=0.48\textwidth]{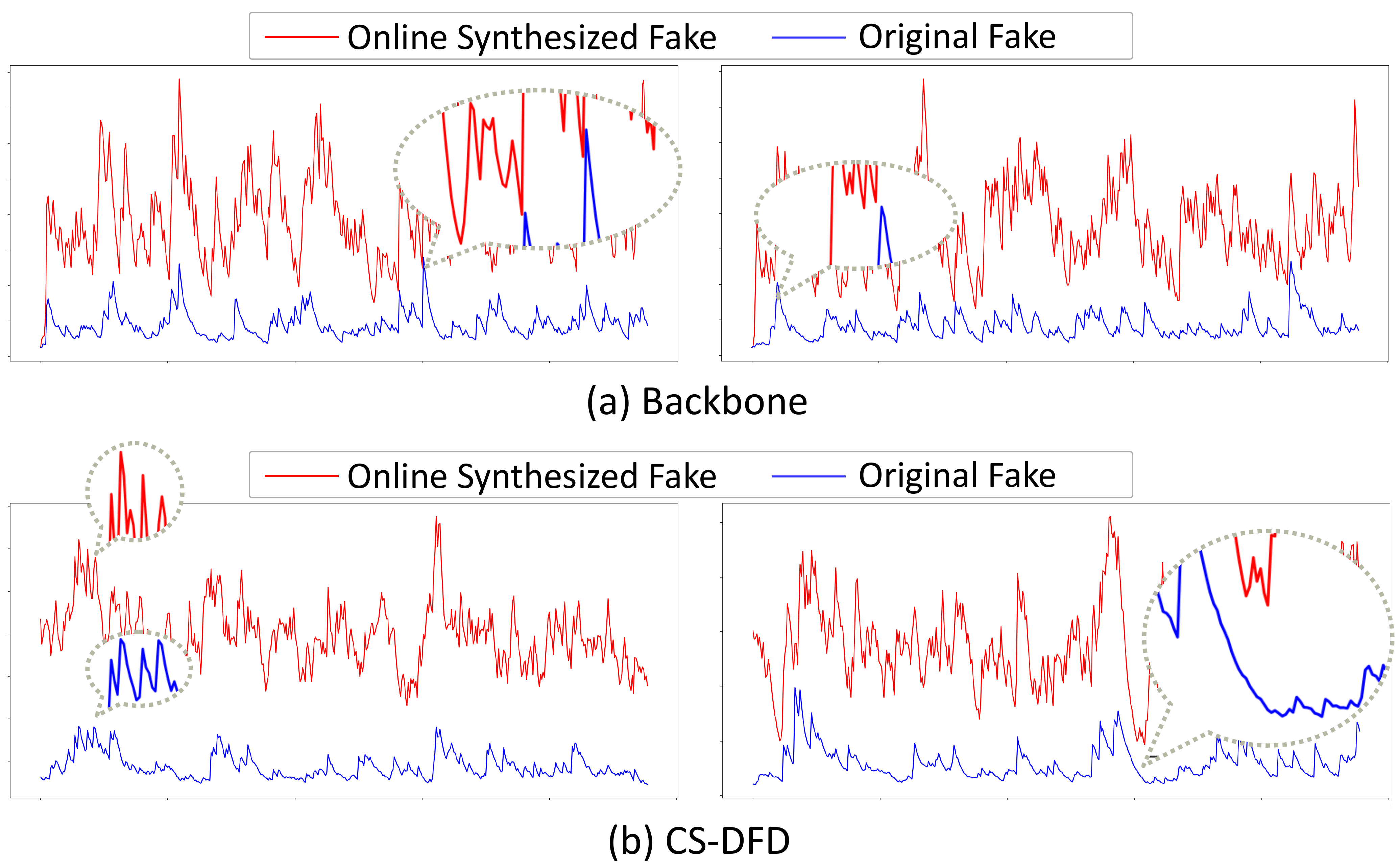}
\caption{Loss trend comparison of different methods when trained simultaneously on original forgeries and online synthesized forgeries. For each model, we present two non-cherry-picked training segments. In each subplot, the $x$-axis is the training steps, and the $y$-axis represents the loss value.}
\label{fig_loss}
\end{figure}
\subsubsection{Loss Comparisons}
In Fig.~\ref{fig_loss}, we illustrate the gradient conflict in the backbone model by analyzing the loss trends before and after applying the CS-DFD method. Specifically, Fig.~\ref{fig_loss}(a) shows the loss curves when the model is trained using the traditional approach with [real, online synthesized images, original fake images]. It can be observed that a strong antagonistic pattern exists between the two types of fake data: when the loss for one type peaks, the loss for the other reaches a minimum. This phenomenon indicates a fundamental conflict in optimization directions induced by the two types of data, which prevents the model from effectively learning meaningful forgery patterns. As a result, the model performs poorly on both in-domain and out-of-domain data. In contrast, our proposed CS-DFD method mitigates this conflict. As shown in the figure, the losses for both types of data follow a consistent and coordinated trend, suggesting that they contribute synergistically to model optimization. This enables the model to maintain strong in-domain detection performance while simultaneously enhancing its generalization capability.

\subsubsection{Gradient Vector Field Analysis}
\begin{figure}[t]
\centering
\includegraphics[width=0.48\textwidth]{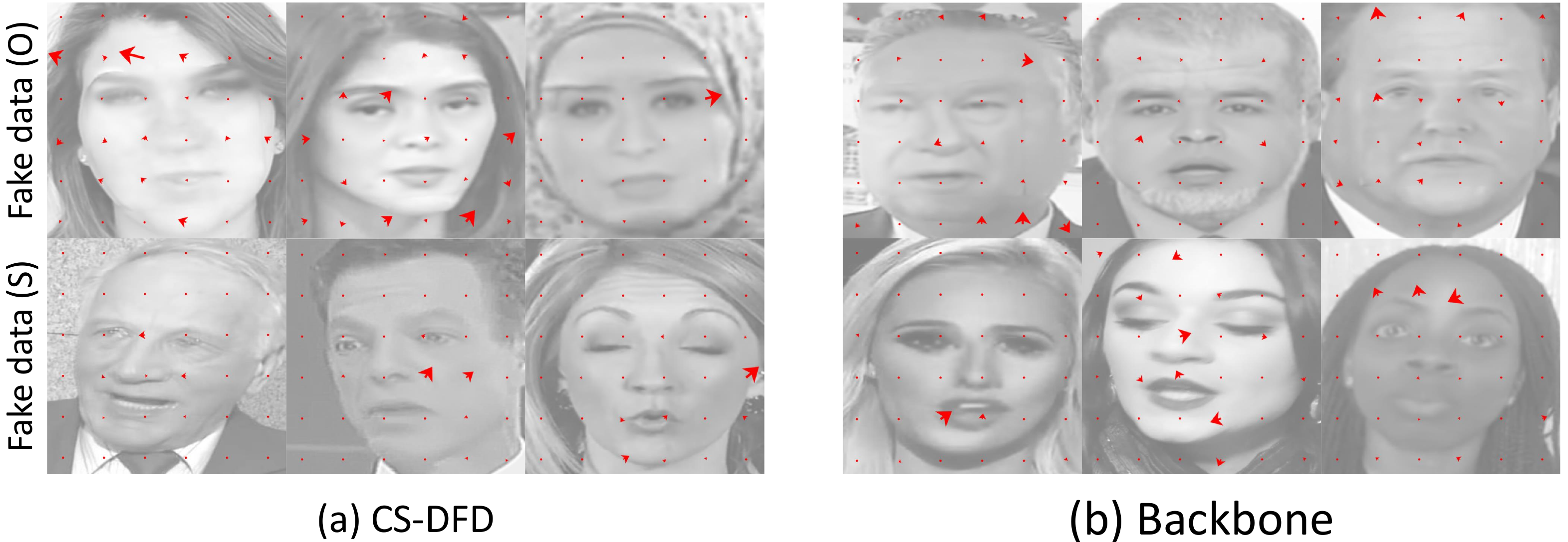}
\caption{Visualization of vector fields at different pixel locations for selected training samples. (a) and (b) present the results of our CS-DFD framework and the backbone model, respectively. It can be observed that, under the guidance of CS-DFD, samples within the same training batch (aligned in the same column) exhibit more consistent gradient directions.}
\label{fig_quiver}
\end{figure}

Fig.~\ref{fig_quiver} illustrates the gradient vector field generated using the `quiver' function for a single sample during training. Fig.~\ref{fig_quiver}(a) and (b) show the results obtained using our proposed model CS-DFD, and the backbone model, respectively. The red arrows evenly distributed across the image pixels represent the local gradients at different positions. The direction indicated by each arrow represents the direction of the gradient, and the length of the arrow corresponds to the magnitude of the gradient. Meanwhile, to enhance the contrast of the visualization results, the original input images are converted to grayscale to display the gradient vectors better. By adjusting the scale factor in the `quiver' function, only the dominant gradients are shown.

To demonstrate the effectiveness of CS-DFD in mitigating gradient conflicts, we arrange both the originally fake images and online synthesized fake images within the same training batch in the same column. As the gradient vector fields show, training with our CS-DFD, samples from different sources in the same batch exhibit more consistent dominant gradient directions, indicating that the inter-source gradient conflicts have been significantly alleviated.

\subsubsection{Hyperparameters Analysis}
\label{sec:hyper}
In the proposed method, two hyperparameters play a critical role: $c$ in Equation~(\ref{equ:neighbor}) and $\tau$ in Equation~(\ref{eq:g*g}). The former controls the degree to which the update vector $v$ deviates from the original gradient $g_0$, ensuring that the new vector remains sufficiently close to preserve training stability while deviating enough to resolve conflicts caused by the original gradient. This enables the optimization to directly benefit from the conflict-free update vector obtained via the UVS module. The latter hyperparameter, $\tau$, governs the sensitivity of the diagonal approximation of the Hessian matrix and directly influences how strongly the conflict loss steers the update direction.

\begin{figure}
    \centering
    \includegraphics[width=0.5\textwidth]{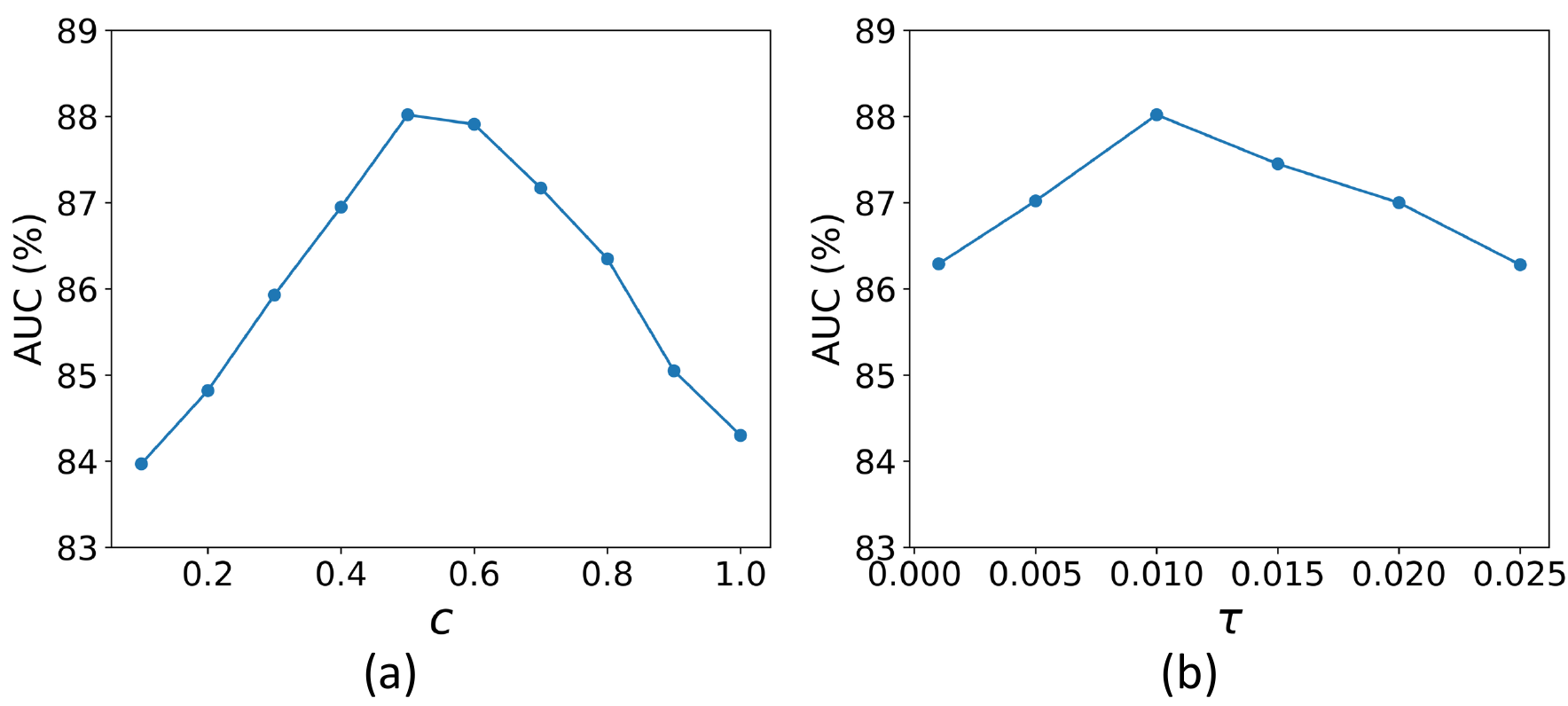}
    \caption{Hyperparameter sensitivity analysis for $c$ and $\tau$. (a) Adjusting $c$ changes the search range of the update vector, which in turn affects model performance. (b) Varying $\tau$ influences the degree of conflict suppression in the embedding space, impacting the generalization ability.}
    \label{fig:hyperparameter}
\end{figure}

We present a detailed hyperparameter analysis in Fig.~\ref{fig:hyperparameter}. The impact of $c$ on model performance exhibits a typical sweet spot curve: as $c$ increases, the AUC on the target domain initially improves, reaches a peak, and then declines. When $c$ is too small, the update vector closely resembles the original gradient, failing to effectively mitigate gradient conflicts. As $c$ increases, these conflicts are alleviated, leading to better generalization. However, excessively large values of $c$ may cause the update direction to deviate too far from the true optimization trajectory, resulting in less reliable result.
A similar trend is observed for $\tau$. An appropriately chosen $\tau$ effectively constrains conflicting update directions without overly suppressing feature diversity, thereby enhancing generalization. In contrast, a large $\tau$ leads to excessive smoothing of projected features, which suppresses expressive capacity and ultimately causes a drop in AUC.
Based on these observations, we empirically set $c = 0.5$ and $\tau = 0.01$ as the default hyperparameter configuration for our experiments.

\subsection{Efficiency Analysis}
In our experiments, we observe that the computational efficiency of the CS-DFD method is nearly identical to that of the original backbone models. 
Over the course of the entire training process, CS-DFD incurs only a modest increase in computational time—approximately 2\% more than the backbone alone (20.3 hours vs. 20 hours), which is acceptable given the significant gains in generalization performance.
A critical reason for this is that both modules in CS-DFD are designed to be lightweight.
Specifically, CS-DFD comprises two modules: UVS and CGR. For UVS, during parameter update, gradients are computed separately for original fakes and online synthesized fakes. An update vector is then constructed to minimize the conflict between these two sources. This process involves computing the norms, inner product, and a linear combination of ${g}_1$ and ${g}_2$, which has a time complexity of $\mathcal{O}(P)$, where $P$ denotes the total number of model parameters. This cost is negligible when compared to a full backward pass.
For CGR, it introduces a small trainable projection layer $\phi_{\theta_p}$ and a conflict descent loss $\mathcal{L}_\phi$, which jointly learn embeddings that mitigate gradient conflict. The output dimension $d$ of $\phi_{\theta_p}$ is much smaller than $P$, making its forward and backward passes computationally inexpensive. The conflict loss employs a Hessian-diagonal approximation (as described in Equation~(\ref{eq:g*g})), which involves only element-wise multiplications over the projected feature vectors and has a time complexity of ${\mathcal{O}}(d)$, and $d \ll P$.
In summary, CS-DFD introduces virtually no additional computational overhead, while effectively addressing gradient conflict through its lightweight and efficient design.

\section{Conclusion}
This work provides an in-depth analysis of the `$1+1<2$' performance degradation phenomenon in deepfake detection, identifying conflicting gradients from heterogeneous forgery data as the primary cause of poor cross-domain generalization. To address this challenge, the Conflict-Suppressed Deepfake Detection (CS-DFD) framework is proposed.
This framework simultaneously aligns the descent directions of losses from diverse data sources and reduces representation-level gradient discrepancies.
Extensive experiments on multiple deepfake benchmarks validate the effectiveness of CS-DFD and confirm that suppressing gradient conflicts is essential for achieving robust, generalizable deepfake detection.


\bibliographystyle{IEEEtran}
\bibliography{grad-tpami}




\vfill

\end{document}